\documentclass[a4paper,conference]{IEEEtran}

\usepackage{graphicx}
\usepackage{comment}
\usepackage{amsmath,amssymb}
\usepackage{cite}
\usepackage{xcolor}
\usepackage{times}
\usepackage{bm}
\usepackage{amsmath}
\newcommand{\mat}[1]{\bm{#1}}
\newcommand{\ten}[1]{\bm{\mathcal{#1}}}
\usepackage{booktabs}
\usepackage{multirow}

\usepackage[flushleft]{threeparttable}
\usepackage{subfigure}
\usepackage{algorithm}
\usepackage{algorithmic}
\usepackage{tabu}
\usepackage{amsthm}
\usepackage{lipsum}
\theoremstyle{definition}
\newtheorem{definition}{Definition}
\usepackage[colorlinks,
            linkcolor=blue,       
            anchorcolor=blue,  
            citecolor=blue,       
            ]{hyperref}

\begin{document}

\title{Coarse to Fine: Image Restoration Boosted by Multi-Scale Low-Rank Tensor Completion}

\author{\IEEEauthorblockN{Rui Lin, Cong Chen, Ngai Wong}
\IEEEauthorblockA{Department of Electrical and Electronic Engineering, 
The University of Hong Kong, Hong Kong\\
Email: \{linrui, chencong, nwong\}@eee.hku.hk}}

\maketitle

\begin{abstract}
Existing low-rank tensor completion (LRTC) approaches aim at restoring a partially observed tensor by imposing a global low-rank constraint on the underlying completed tensor. However, such a global rank assumption suffers the trade-off between restoring the originally details-lacking parts and neglecting the potentially complex objects, making the completion performance unsatisfactory on both sides. To address this problem, we propose a novel and practical strategy for image restoration that restores the partially observed tensor in a coarse-to-fine (C2F) manner, which gets rid of such trade-off by searching proper local ranks for both low- and high-rank parts. Extensive experiments are conducted to demonstrate the superiority of the proposed C2F scheme. The codes are available at: \href{https://github.com/RuiLin0212/C2FLRTC}{https://github.com/RuiLin0212/C2FLRTC}.
\end{abstract}

\IEEEpeerreviewmaketitle


\section{Introduction}
\label{sec:intro}
In the past decade, computer vision and image processing have received immense attention due to the proliferation of deep learning and artificial intelligence (AI). Image restoration is one of the vital research topics in this field as the need for large amounts of high quality images is intrinsic to various AI tasks. Image restoration aims to recover the missing pixels in an image based on only partially observed data, thus forming a clearer image. Therefore, deciphering the relations between the known and unknown entries is a key challenge in image restoration tasks.

Existing approaches for image restoration can generally be divided into four categories~\cite{bertalmio2000image}, namely,  partial differential equations (PDEs)-based diffusion, examplar-based techniques, hybrid algorithms, and low-rank tensor completion (LRTC) methods. PDEs-based methods~\cite{voci2004estimating,telea2004image,kumar2019linear} solve the restoration problem by using smoothness priors to diffuse local structures from the known region to the missing unknown region. Whereas the examplar-based techniques~\cite{aujol2010exemplar,le2011examplar} restore the missing parts by sampling, copying, or stitching together the observed parts of the image. Since PDEs-based diffusion works well when completing missing areas like straight lines and curves but shows cons in texture recovering that examplar-based techniques are good at, hybrid algorithms~\cite{starck2005image,bugeau2010comprehensive} are proposed to make the first two categories complementary so as to restore texture while preserving edges or structures. We remark that all these aforementioned schemes require searches to capture the local correlation or dependencies among the pixels. Different from them, LRTC methods~\cite{liu2012tensor,yokota2017simultaneous,hauenstein2019homotopy} aim to dig out the statistical properties of the image and capture the global structure of the input data. 

\begin{figure}[t]
\centering
\includegraphics[scale=0.36]{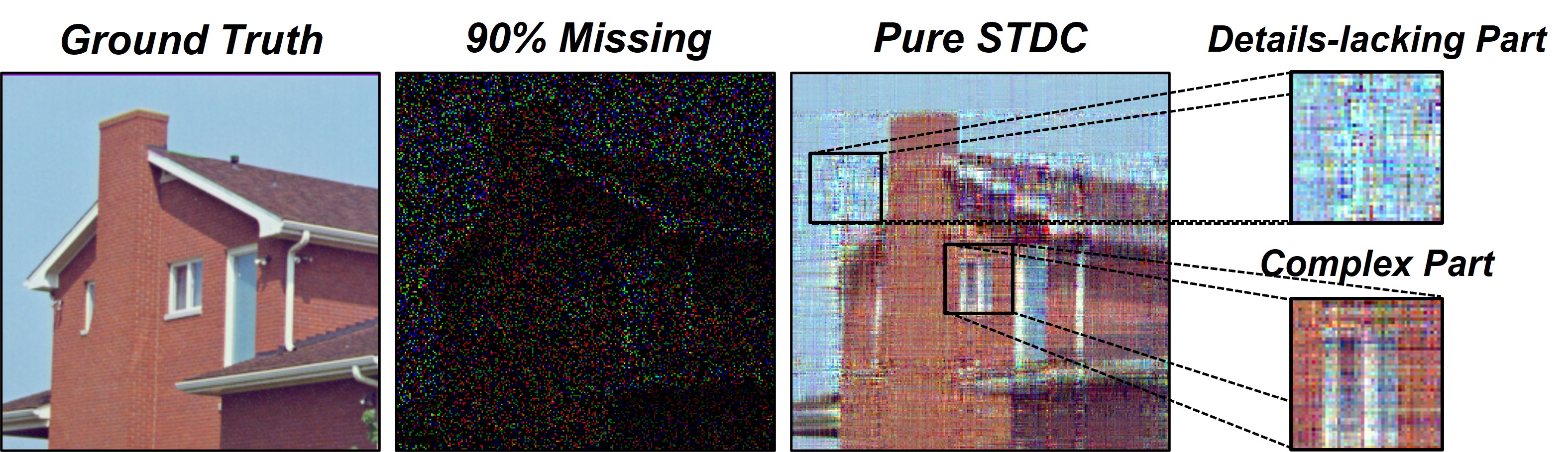}
\vspace{-2mm}
\caption{There is a global low-rank assumption in most LRTC methods, which can be regarded as a trade-off between restoring the originally low-rank parts and neglecting the potentially high-rank complex objects. For easy understanding and better visualization, we use STDC~\cite{chen2013simultaneous} as an example. Under the global low-rank assumption, it is seen that the restored sky is not smooth while the window loses the details.}
\vspace{-4mm}
\label{fig:motivation}
\end{figure}

Although LRTC breaks through the limitation of relying only on the observed adjacent pixels, and adopts a global low-rank assumption on the completed image, we find the global low-rank setting embodies a \textbf{\textit{trade-off}} between restoring the originally low-rank parts and neglecting the potentially high-rank parts. The trade-off, depicted in Fig.~\ref{fig:motivation}, brings about two issues: 1) the restoration of the parts associated with a lower local rank will be hindered with insufficient observed pixels, 2) the potentially high-rank complex objects will suffer the loss of details due to over-smoothing. Therefore, the performance of LRTC can potentially be enhanced by preserving the benefits brought by the global rank setting, while catering for local structures simultaneously. 

To this end, we propose a novel multi-scale LRTC strategy that restores the partially observed image in a \textbf{C}oarse-to-\textbf{F}ine (C2F) manner. Succinctly, the image is first completed in the coarse stage, where the whole image is restored by assuming a global low-rank tensor structure. After capturing the overall data structure in the first stage, the partially observed image is divided into smaller patches in the subsequent fine stage, where the patches are completed independently using the same LRTC method. The restored patches then replace their counterparts in the coarse completed image if they fulfill a designated metric. Noticeably, the fine-grained stage can be repeated multiple times to seek a satisfactory restored image, and the experimental results demonstrate the necessity of such \textbf{\textit{successive}} fine-grained completion. We highlight that the performance of existing LRTC methods can be boosted taking advantage of this \textbf{\textit{divide-and-conquer}} strategy. Our main contributions are: 
\begin{itemize}

\item A general and intuitive C2F strategy is proposed, which effectively enhances the performance of existing LRTC methods by seeking proper local ranks for both the low- and high-rank parts, respectively. 

\item No more trade-off that aims to balance the restoration performance of simple and complex parts is required. 

\item Utilization of the data from both coarse and fine hierarchies, thus capturing both the global and local data structures simultaneously.

\item  Extensive experiments and ablation study for validating C2F, which demonstrate the superiority of C2F in image completion versus existing algorithms.
\end{itemize}


\section{Related Works}
\label{sec:lit_review}
The methods to determine the global low-rank constraints in the LRTC optimization problem can generally be divided into two categories, namely, by generalizing the low-rank matrix completion techniques or by employing tensor decomposition. 

For the first category, Liu et al.~\cite{liuiccv} are the pioneers that proposed tensor trace norm and defined it as the average of trace norms of all unfolded matrices. The tensor with missing entries is restored by minimizing this metric. However, solving the optimization problem is nontrivial due to the inter-dependency between the unfolded matrices. Therefore, two enhanced methods (viz. FaLRTC and HaLRTC) are proposed~\cite{liu2012tensor} wherein a relaxation is employed to separate the dependent relations between the unfolded matrices such that a globally optimal solution is achieved. However, implementing the singular value decomposition needed in the tensor trace norm is computationally expensive. To alleviate the computational burden, a parallel matrix factorization approach~\cite{xu2013parallel} is proposed, which expresses each unfolded matrix of the underlying tensor as the product of two low-rank factors that are alternately updated.

The second category employs tensor decomposition techniques. For example, Canonical Polyadic (CP) decomposition~\cite{kolda2009tensor} is applied on the partially observed tensor in~\cite{tomasi2005parafac,acar2011scalable}, which predicts the missing entries through the low-rank constraint. In~\cite{imaizumi2017tensor}, the underlying tensor structure is assumed to be a low-rank Tensor Train (TT)~\cite{oseledets2011tensor}, whose TT-cores are updated by solving the alternating least squares problem. However, manually specifying the suitable tensor ranks for these methods is challenging. To avoid that, the authors of~\cite{liu2014factor} and~\cite{liu2014generalized} applied trace norm regularization to the factor matrices of CP and Tucker decomposition~\cite{kolda2009tensor}, respectively. The low-rank constraint is then achieved by adjusting the weight of trace norm regularization term in the optimization problem.

Recently, Xue et al.~\cite{xue2021multilayer} proposed a completion approach using multilayer sparsity-based tensor decomposition, which further decomposes the CP factors several times. We remark that their algorithm still has the global low-rank assumption for the first CP decomposition, on which all the following decomposition is based. In contrast, the fine-grained completion in our strategy is not entirely dependent on the coarse stage, which instead allows the LRTC methods to correct the global assumption if the ranks are improper in the former stages. Multiscale feature (MSF) is proposed in~\cite{zhang2021multiscale}, which aims to build a higher-order tensor containing more image information, and then applies TT decomposition on it. In fact, MSF-TT is orthogonal to the proposed C2F strategy, whose performance can be further improved by the latter. 

In short, the performance of existing LRTC methods belonging to \textit{\textbf{either}} category can be enhanced with C2F, which is able to exploit both the global and local tensor rank information, allowing a more comprehensive utilization of the partially observed data.

\section{Preliminary}
\label{sec:preliminary}
Tensors are multi-mode arrays, and a $d$-way tensor is denoted as $\ten{A}\in\mathbb{R}^{I_1\times I_2 \times \cdots \times I_d}$. The element of $\ten{A}$ is represented by $\ten{A}(i_1,i_2,\ldots,i_d)$, where 1$\le$ $i_k$ $\le$ $I_k$, $k = 1,2,\ldots,d$. The numbers $I_1, I_2, \ldots, I_d$ are called the dimensions of the tensor $\ten{A}$.
We use boldface capital calligraphic letters $\ten{A}$, $\ten{B}$, $\ldots$ to denote tensors, italic boldface capital letters $\mat{A}$, $\mat{B}$, $\ldots$ to denote matrices, italic boldface letters $\mat{a}$, $\mat{b}$, $\ldots$ to denote vectors, and roman letters $a$, $b$, $\ldots$ to denote scalars. 

\begin{definition}\textit{(Tensor mode-$k$ matricization~\cite{kolda2009tensor}, p. 459)} The mode-$k$ matricization of a tensor $\ten{A}\in\mathbb{R}^{I_1\times\cdots \times I_k\times\cdots\times I_d}$ denoted by $\mat{A}_{(k)}\in\mathbb{R}^{I_k\times (I_1 \times \cdots \times I_{k-1}\times I_{k+1}\times\cdots\times I_d)}$ and tensor element $\ten{A}(i_1,i_2,\cdots,i_d)$ maps to matrix element $\mat{A}_{(k)}(i_k,j)$, where 
\begin{align}
j=1+\sum\limits_{n=1,n\neq k}^{d}(i_n-1)J_n \quad \text{with} \quad J_n=\prod\limits_{m=1,m\neq k}^{n-1}I_m.
\end{align}
\end{definition}

\begin{definition}\textit{(Tensor mode-k product~\cite{kolda2009tensor}, p. 460)} Tensor mode-$k$ product involves a multiplication of a matrix with a $d$-way tensor along one of its $d$ modes. The mode-$k$ product $\ten{B}=\ten{A}\, {\times_k}\, \bm{U}$ of a tensor $\ten{A}\in\mathbb{R}^{I_1\times\cdots \times I_k\times\cdots\times I_d}$ with a matrix $\bm{U}\in\mathbb{R}^{R\times I_k}$ is defined by%
\begin{equation}
\small{
\begin{aligned}
&\ten{B}(i_1,\ldots,i_{k-1},j,i_{k+1},\ldots,i_d)\\=&\sum\limits_{i_k=1}^{I_k}  \mat{U}(j,i_k)\, \ten{A}(i_1,\ldots, i_{k-1},i_k,i_{k+1},\ldots,i_d),%
\end{aligned}%
}
\end{equation}
where $\ten{B}\in\mathbb{R}^{I_1\times\cdots \times I_{k-1}\times R \times I_{k+1}\times\cdots\times I_d}$.
\end{definition}

\begin{definition}\textit{(Tensor Frobenius norm)} The Frobenius norm of a tensor $\ten{A}\in\mathbb{R}^{I_1\times I_2 \times \cdots\times I_d}$ is defined as
\begin{align}
||\ten{A}||_F=\sqrt{\sum\limits_{i_1=1}^{I_1} \sum\limits_{i_2=1}^{I_2}\cdots \sum\limits_{i_d=1}^{I_d} \ten{A}(i_1,i_2,\cdots,i_d)^2}.
\end{align}
\end{definition}

\begin{definition}\textit{(Tucker decomposition)} The Tucker decomposition decomposes a $d$-way tensor $\ten{A} \in \mathbb{R}^{I_1 \times \cdots \times I_d}$ into a smaller Tucker core $\ten{S} \in \mathbb{R}^{R_1 \times \cdots \times R_d}$ and orthogonal factor matrices $\mat{U}_i \in \mathbb{R}^{I_i \times R_i}\,(i=1,\ldots,d)$ as
\begin{align}
\ten{A}&= \ten{S} \times_1 \mat{U}_1 \times_2 \cdots \times_d \mat{U}_d.
\label{eq:HOSVD}
\end{align}
\end{definition}

\begin{figure*}[t]
\centering
\includegraphics[scale=1]{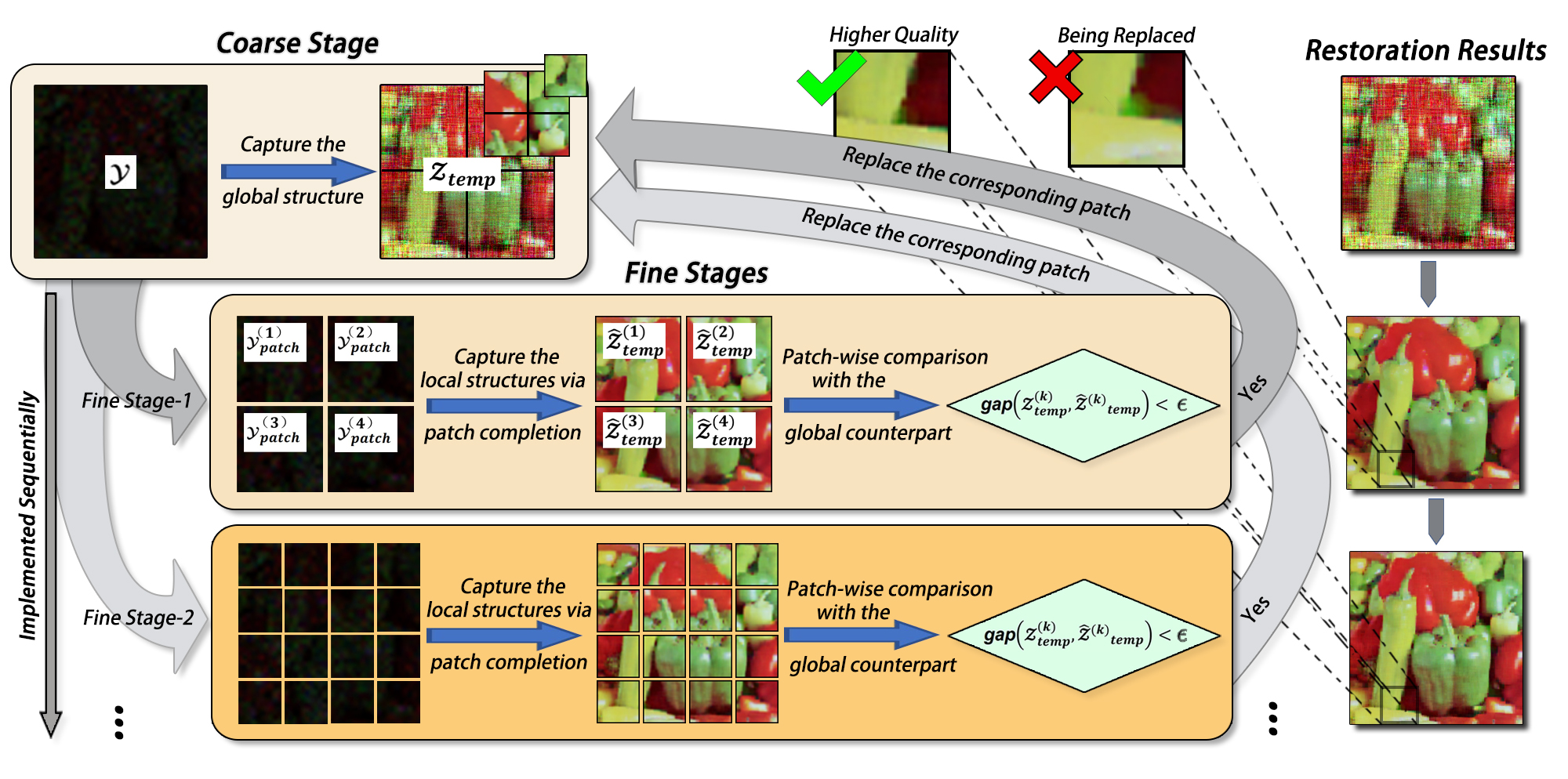}
\vspace{-2mm}
\caption{The workflow of the C2F strategy. First, a coarse completed image $\ten{Z}_{temp}$ is obtained. Then, for each fine stage, the original partially observed image $\ten{Y}$ is separated into $N$ same-size patches, namely $\ten{Y}^{(1)}_{patch}$, $\ten{Y}^{(2)}_{patch}$, $\cdots$, $\ten{Y}^{(N)}_{patch}$. Each patch has overlapping pixels with its neighbors to achieve a smooth patch boundary during subsequent completion. Those patches are then completed separately via LRTC method, and $N$ fine-grained completed patches can be obtained, namely, $\hat{\ten{Z}}_{temp}^{(1)}$, $\cdots$, $\hat{\ten{Z}}_{temp}^{(N)}$. The newly obtained patches satisfying the criterion will replace the corresponding patches in the $\ten{Z}_{temp}$ image. The number of fine stages can be set according to the restoration requirements.}
\vspace{-4mm}
\label{fig:flowchat}
\end{figure*}

\section{Low-rank Tensor completion boosted by Coarse to Fine strategy}
\label{sec:c2flrtc}

For easy understanding, we now exemplify C2F using a representative LRTC method called LRTC-TV-II~\cite{li2017low}, which is based on Tucker decomposition. In Section~\ref{subsec:problem}, we first introduce the problem formulation. In Section~\ref{subsec:C2F}, we elaborate the details of the proposed scheme and show the C2F algorithm. Finally, we discuss the necessity of the successive fine-grained completion in Section~\ref{subsec:discuss}.

\subsection{Problem Formulation}
\label{subsec:problem}

Given the partially observed tensor $\ten{Y}\in \mathbb{R}^{I_1\times I_2 \times \cdots \times I_d}$, and the observed tensor entry set $\Omega$, the image restoration problem solved by LRTC-TV-II~\cite{li2017low} is as follows:
\begin{align}
\label{eq:lrtctv}
\nonumber \min_{\ten{Z},\ten{S},\{\mat{U}_k\}_{k=1}^d}         &\quad \, \lambda_1\sum\limits_{k=1}^{d}\beta_k|\mat{F}_k\ten{Z}_{(k)}| + \lambda_2\sum\limits_{k=1}^{d} || \mat{U}_k||_* +\lambda_3||\ten{S}||_F^2\\
\nonumber \textrm{subject to}          &\quad \, \ten{Z}= \ten{S} \times_1 \mat{U}_1 \times_2 \cdots \times_d \mat{U}_d,\\
&\quad [\ten{Z}]_{\Omega}~=~[\ten{Y}]_{\Omega},
\end{align}
where $\ten{Z}$ is the restored tensor, $\ten{Z}_{(k)}$ is the mode-$k$ matricization of $\ten{Z}$. The binary matrix $\mat{F}_k \in \mathbb{R}^{(I_k -1)\times I_k}$ has the entries $\mat{F}_k(i,i) = 1$ and $\mat{F}_k(i,i+1)=-1$, while the remaining elements are all zeros. The term $|\mat{F}_k\ten{Z}_{(k)}|$ represents the total variation (TV) regularization and is computed by summing over the absolute values of all entries in $\mat{F}_k \ten{Z}_{(k)}$. The operator $||\cdot||_*$ denotes the trace norm of a matrix. The completed tensor $\ten{Z}$ is assumed to be in a low-rank Tucker format, namely, $\ten{Z} = \ten{S}\times_1\mat{U}_1\times_2\cdots\times_d \mat{U}_d$, where $\ten{S} \in \mathbb{R}^{I_1\times I_2 \times \cdots \times I_d}$ and $\mat{U}_k \in \mathbf{R}^{I_k \times I_k}$. Note that instead of setting low Tucker ranks directly, the objective~(\ref{eq:lrtctv}) employs a trace norm term to enforce the low-rank constraint in the factor matrix $U_k$. The last term in~(\ref{eq:lrtctv}) avoids overfitting, and $\beta_k$ is either $0$ or $1$ to indicate whether TV normalization is imposed on mode-$k$ of $\ten{Z}$. For color image completion tasks, the setting would be $\beta_1 = \beta_2 = 1$ and $\beta_3 = 0 $, since only spatial modes are expected to be smooth. The above optimization problem can be efficiently solved by alternating direction method of multipliers (ADMM)~\cite{boyd2011distributed}, which introduces auxiliary matrices to separate the inter-dependency between the matrices of~(\ref{eq:lrtctv}).

\vspace{2mm}
\begin{algorithm}[h]
\renewcommand{\algorithmicrequire}{\textbf{Input:}}
\renewcommand{\algorithmicensure}{\textbf{Output:}}
\caption{C2F Strategy with LRTC-TV-II~\cite{li2017low}}
\label{alg:c2flrtc}
\begin{algorithmic}[1]
\REQUIRE Partially observed image $\ten{Y}$; observed image entry set $\Omega$; patch replace threshold $\epsilon$; number of fine stage implementations $F$; weight parameters $\lambda_k$ $(k=1,2,3)$.
\ENSURE The restored image $\ten{Z}$.
\vspace{1ex}
\STATE $\ten{Z}_{temp}$ $\leftarrow$ Coarse stage completion;
\STATE $f=1$ $\leftarrow$ Count the number of fine stages;
\WHILE {$f\leq F$}
\STATE $\ten{Y}_{patch}^{(k)}, k=1,\ldots,2^{2f}$ $\leftarrow$ Divide $\ten{Y}$ into $2^{2f}$ patches of the same size; 
\STATE $\lambda_2 = \mu \lambda_2$, $\mu>1$ $\leftarrow$ Adjust $\lambda_2$ to impose a local rank assumption;
\STATE $\ten{Z}_{temp}^{(k)}, k=1,\ldots,2^{2f}$. $\leftarrow$ Divide $\ten{Z}_{temp}$ according to the same partition used for $\ten{Y}$ in step $4$;
\FOR{ $k$ from $1$ to $2^{2f}$}
\STATE $\hat{\ten{Z}}_{temp}^{(k)}$ $\leftarrow$ Complete the $k$-th patch;
\IF {$\rm{gap}$($\ten{Z}_{temp}^{(k)}$, $\hat{\ten{Z}}_{temp}^{(k)}$) $<  \epsilon$}
\STATE $\ten{Z}_{temp}^{(k)}$ = $\hat{\ten{Z}}_{temp}^{(k)}$ $\leftarrow$ Replace its counterpart;
\ENDIF
\ENDFOR    
\STATE $f=f+1$ $\leftarrow$ Update the number of fine stages;
\STATE Update the threshold $\epsilon$ (cf. (\ref{eq:epsilon}));
\ENDWHILE
\STATE $\ten{Z}$=$\ten{Z}_{temp}$.
\end{algorithmic}%
\end{algorithm}%

\subsection{C2F Strategy}
\label{subsec:C2F}

As shown in Fig.~\ref{fig:motivation}, the existing LRTC methods aim to strike a balance between the potentially low- and high-rank parts, which is achieved by imposing a global low-rank assumption. However, our proposed C2F strategy gets rid of such trade-off by utilizing the data from both coarse and fine hierarchies, and focusing more on the local structure (small patches) in a successive manner. Here we present the details of the C2F strategy, which generally consists of two tensor completion stages, namely, the coarse stage and the fine stage.

\subsubsection{Coarse Stage} In the coarse stage, the whole observed image $\ten{Y}$ is completed by solving the optimization problem~(\ref{eq:lrtctv}) with LRTC-TV-II, and a coarsely completed image~$\ten{Z}_{temp}$ is obtained. Considering the limits brought by the global low-rank assumption on restoring both the details-lacking (low-rank) parts and the complex (high-rank) parts, we proceed to the fine stage to pursue a higher-quality completion result.

\subsubsection{Fine Stage} The fine stage aims to complete the image from a fine-grained viewpoint, which can be repeated multiple times until a satisfactory restored image is achieved. Every single fine stage comprises three steps, namely, fine-grained completion, patch comparison, followed by replacement. 

\textbf{Step 1: Fine-grained Completion.} The original partially observed image $\ten{Y}$ is separated into $N$ patches with the same size, namely $\ten{Y}^{(1)}_{patch}$, $\ten{Y}^{(2)}_{patch}$, $\cdots$, $\ten{Y}^{(N)}_{patch}$. To obtain a smooth boundary during the subsequent completion process, we let every single patch have \textbf{\textit{overlapping}} pixels with its \textbf{\textit{neighbors}}. Those patches are then completed separately with the selected LRTC-TV-II technique with a local rank assumption on them, which is achieved by adjusting $\lambda_2$ in~(\ref{eq:lrtctv}). The local rank is smaller than the global rank we set at the very beginning in the coarse stage and this setting is validated in Section~\ref{exp:local_rank}.

\textbf{Step 2: Patch Comparison.} After the first fine stage completion, $N$ completed patches, $\hat{\ten{Z}}^{(1)}_{temp}$, $\hat{\ten{Z}}^{(2)}_{temp}$, $\cdots$, $\hat{\ten{Z}}^{(N)}_{temp}$, under local rank assumption are obtained. To compare those patches at the same position in different completion stages, we divide the temporarily restored image $\ten{Z}_{temp}$ into $N$ patches following the partitions employed in the current fine stage to obtain $\ten{Z}^{(1)}_{temp}$, $\ten{Z}^{(2)}_{temp}$, $\cdots$, $\ten{Z}^{(N)}_{temp}$. We use Relative Squared Error (RSE) to quantitatively compare the difference between the two corresponding patches, which is formulated as below:
\begin{align}
{\rm gap}(\ten{\hat{Z}}_{temp}^{(k)},\ten{Z}_{temp}^{(k)}) = \frac{||\ten{\hat{Z}}^{(k)}_{temp}-\ten{Z}_{temp}^{(k)}||_F}{||\ten{Z}_{temp}^{(k)}||_F}.
\end{align}
The greater the value, the greater the difference between the completion results of the selected patch in the different stages.

\textbf{Step 3: Replacement.} Based on the difference between the two corresponding patches, we rely on a pre-defined threshold $\epsilon$ to decide whether to replace $\ten{Z}_{temp}^{(k)}$ with $\ten{\hat{Z}}_{temp}^{(k)}$:
\begin{equation}
\ten{Z}_{temp}^{(k)}=\left\{
\begin{array}{l}
    \ten{Z}_{temp}^{(k)}\text{, if ${\rm gap}(\ten{\hat{Z}}_{temp}^{(k)},\ten{Z}_{temp}^{(k)})\geq \epsilon$,} \\
    \ten{\hat{Z}}_{temp}^{(k)}\text{, if ${\rm gap}(\ten{\hat{Z}}_{temp}^{(k)},\ten{Z}_{temp}^{(k)})< \epsilon$.} \\
\end{array}
\right.
\end{equation}

The reason to apply this replacement scheme is that the completed image in the coarse stage captures the overall structure of the underlying ground truth, if the newly completed patch $\ten{\hat{Z}}_{temp}^{(k)}$ has a large gap between its counterpart in $\ten{Z}_{temp}$, it risks deviating from the ground truth severely. Therefore, we only replace those eligible newly completed patches into the corresponding positions of $\ten{Z}_{temp}$, where the overlapping region of patches takes the averaged pixel values.

Since the assumed local rank is not necessarily suitable for the patches in the fine stage, we prefer to implement the fine stage several times in a sequential manner. The size of the separated patch decreases along with the sequence of fine stages. By repeating the fine stages, both the complex and details-lacking parts are more likely to meet the suitable tensor rank benefiting from the simultaneously reduced patch size and assumed local rank in a divide-and-conquer manner. The whole procedure of C2F strategy with specific LRTC-TV-II is summarized in Algorithm~\ref{alg:c2flrtc}, and the detailed workflow of the \textbf{\textit{general}} C2F strategy that can be combined with any LRTC methods is displayed in Fig.~\ref{fig:flowchat}.

\subsection{Necessity of Successive Fine-grained Completion}
\label{subsec:discuss}

The earlier fine stages with relatively large patches are expected to capture the sub-global structure of the whole image. If all intermediate fine stages are dropped, the transition from global to local viewpoint will be too rapid, which ignores the important sub-global relations between patches and weight too much on the fine-grained neighbors. The omission leads to restored results that are far from the ground truth. In Section~\ref{exp:gradual_refine}, experimental results are provided which compare the proposed C2F  with the Short-Cut C2F (which contains only the coarse stage at the beginning and the last fine stage with the smallest patches) to verify this statement.

\section{Experimental Results}
\label{sec:experiments}
\begin{figure}
\centering
\includegraphics[scale=0.36]{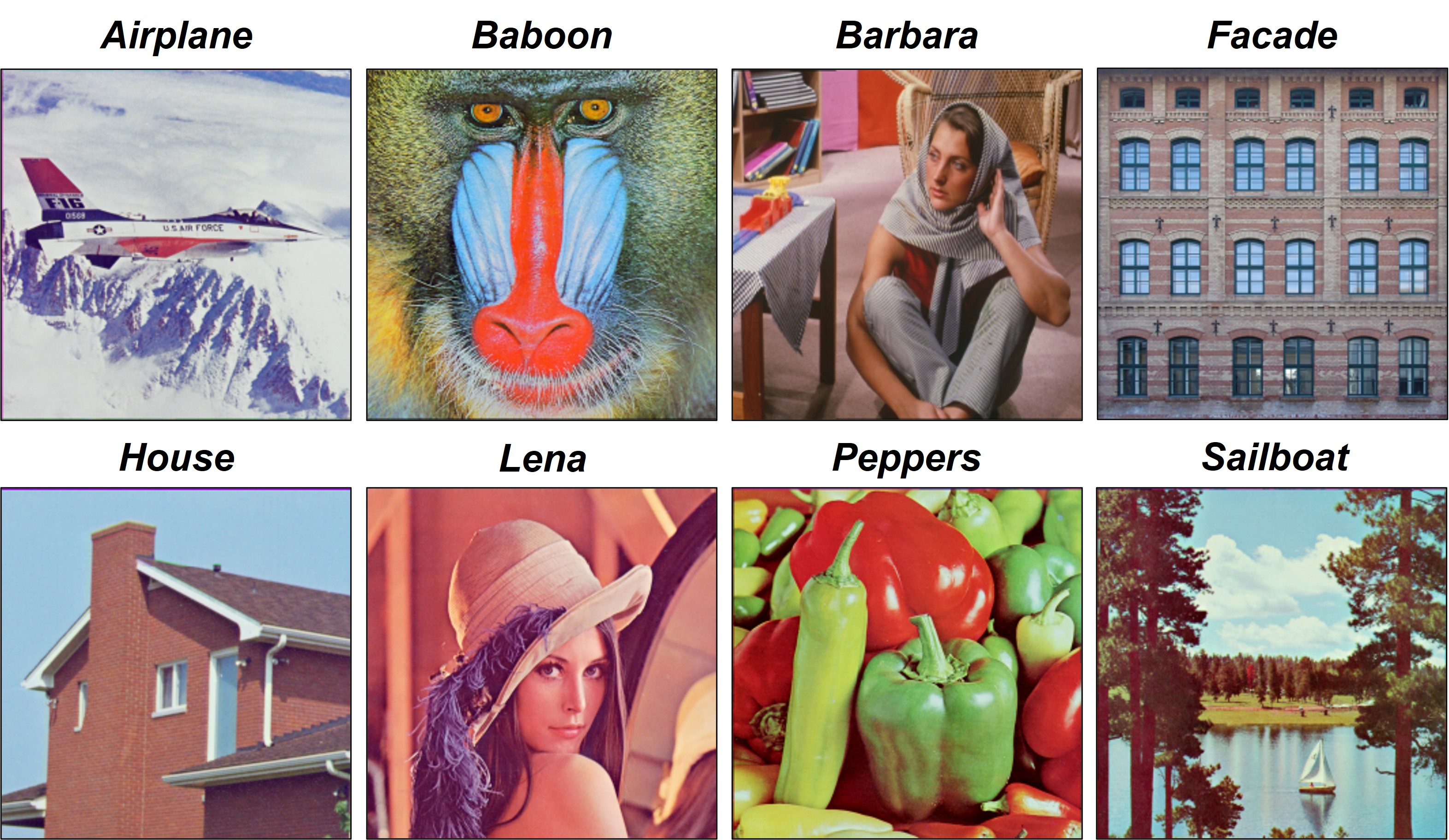}
\vspace{-3mm}
\caption{Ground truth of the eight benchmark color images.}
\label{fig:groundtruth}
\end{figure}

Extensive numerical experiments are conducted to evaluate the color image restoration performance of the pure LRTC methods and different versions of C2F strategy. In Section~\ref{exp:pure_vs_C2F}, we select four LRTC algorithms with publicly available official implementations, namely, HaLRTC$^1$\cite{liu2012tensor}\footnote{$^1$The code of HaLRTC is available at: \href{https://github.com/andrewssobral/mctc4bmi/blob/master/algs_tc/LRTC/HaLRTC.m}{https://github.com/halrtc}}, STDC$^2$\cite{chen2013simultaneous}\footnote{$^2$The code of STDC is available at: \href{http://mp.cs.nthu.edu.tw/project_STDC}{http://mp.cs.nthu.edu.tw/project\_stdc}}, LRTC-TV-II$^3$\cite{li2017low}\footnote{$^3$The code of LRTC-TV-II is available at: \href{https://xutaoli.weebly.com/}{https://xutaoli.weebly.com}}, and LRTC-PDS$^4$\cite{yokota2017simultaneous}~\footnote{$^4$The code of LRTC-PDS is available at: \href{https://drive.google.com/file/d/1o6pFTQFPX_2eSvXotF9KnHLanSTe7Set/view}{https://drive.google.com/lrtc\_pds}}, comparing their performance with and without the proposed C2F strategy. Fig.~\ref{fig:groundtruth} shows the ground truth of the eight benchmark color images, and the size of each image is $256 \times 256 \times 3$. In Section~\ref{exp:gradual_refine}, we compare the proposed C2F with its short-cut version to demonstrate the effectiveness of the gradual refinement. Additionally, in Section~\ref{exp:local_rank}, we demonstrate the rationality of the decreasing local rank setting employed in our experiments. All computations were done on an Intel(R) Core(TM) i5-6500 processor running at 3.2GHz with 16GB RAM, and the implementation platform is MATLAB 2020b.

\begin{figure}
\vspace{-2mm}
\centering
\includegraphics[scale=0.36]{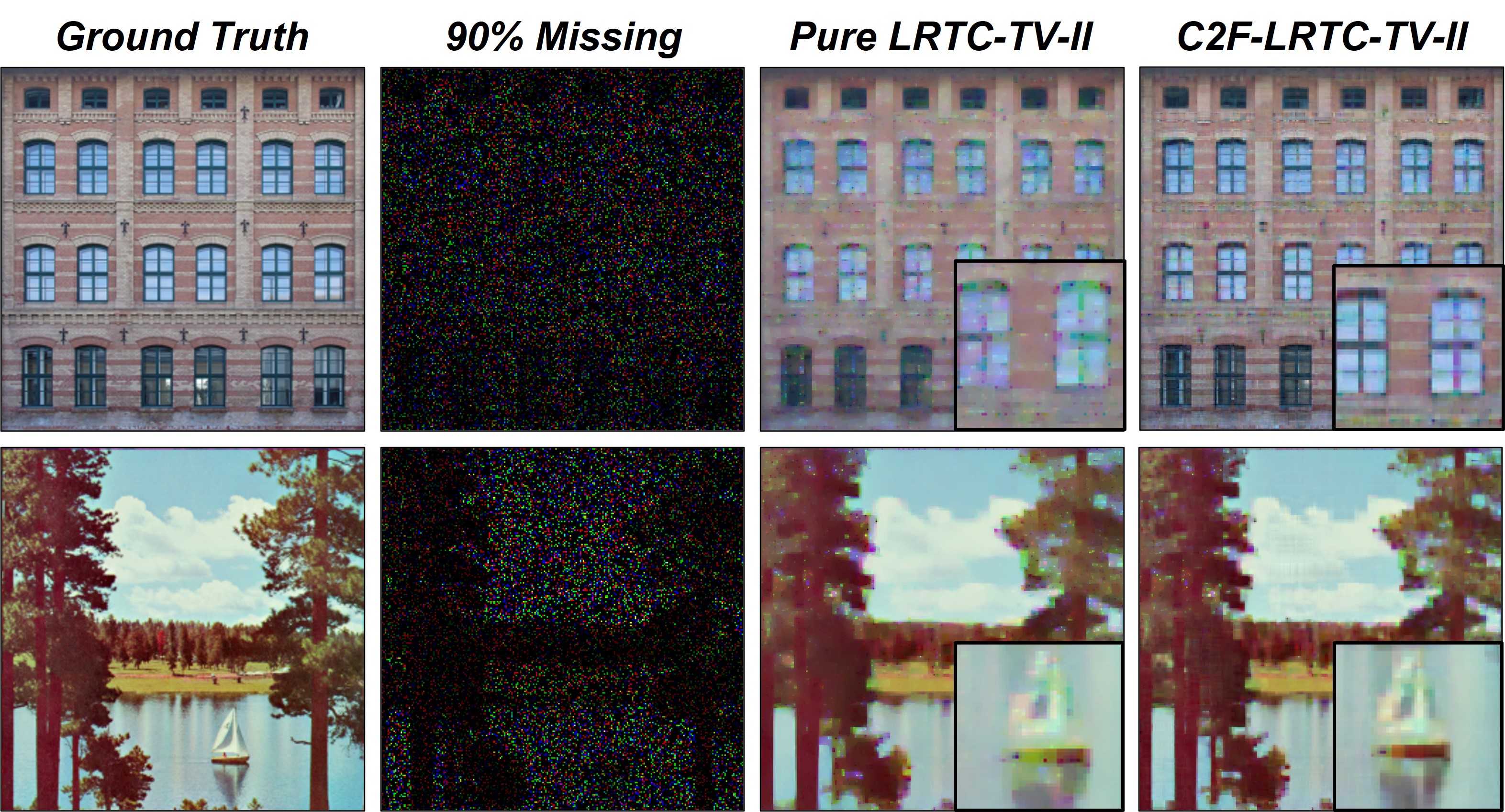}
\vspace{-3mm}
\caption{The C2F strategy helps LRTC methods to restore more details.}
\label{fig:facade_example}
\vspace{-4mm}
\end{figure}

\begin{table}[t]
\scriptsize
\centering
\caption{Performance comparison between three different completion strategies combined with LRTC-TV-II~\cite{li2017low}. The missing ratio of the eight benchmark color images is all $90\%$.}
\label{tab:short_cut}
\setlength{\tabcolsep}{0.7mm}{
\begin{tabular}{lcccccccc}
\toprule
\multirow{2}{*}{\footnotesize{\textbf{Method}}} & \multicolumn{2}{c}{\footnotesize{\textbf{Airplane}}}  & \multicolumn{2}{c}{\footnotesize{\textbf{Baboon}}} & \multicolumn{2}{c}{\footnotesize{\textbf{Barbara}}} & \multicolumn{2}{c}{\footnotesize{\textbf{Facade}}} \\
~ & \textbf{PSNR} & \textbf{RSE} & \textbf{PSNR} & \textbf{RSE} & \textbf{PSNR} & \textbf{RSE} & \textbf{PSNR} & \textbf{RSE} \\
\midrule
LRTC-TV-II & $22.80$ & $0.098$ & $21.22$ & $0.165$ & $23.87$ & $0.131$ & $21.79$ & $0.158$\\
C2F-LRTC-TV-II & $\textcolor{blue}{\mathbf{23.28}}$ & $\textcolor{blue}{\mathbf{0.085}}$ & $\textcolor{blue}{\mathbf{23.05}}$ & $\textcolor{blue}{\mathbf{0.161}}$ & $\textcolor{blue}{\mathbf{24.75}}$ & $\textcolor{blue}{\mathbf{0.119}}$ & $\textcolor{blue}{\mathbf{24.40}}$ & $\textcolor{blue}{\mathbf{0.117}}$ \\
Short-Cut C2F & $22.77$ & $0.093$ & $19.64$ & $0.193$ & $20.51$ & $0.193$ & $21.07$ & $0.171$ \\
\midrule
\midrule
\multirow{2}{*}{\footnotesize{\textbf{Method}}} & \multicolumn{2}{c}{\footnotesize{\textbf{House}}}  & \multicolumn{2}{c}{\footnotesize{\textbf{Lena}}} & \multicolumn{2}{c}{\footnotesize{\textbf{Peppers}}} & \multicolumn{2}{c}{\footnotesize{\textbf{Sailboat}}} \\
~ & \textbf{PSNR} & \textbf{RSE} & \textbf{PSNR} & \textbf{RSE} & \textbf{PSNR} & \textbf{RSE} & \textbf{PSNR} & \textbf{RSE} \\
\midrule
LRTC-TV-II & $24.99$ & $0.094$ & $25.36$ & $0.097$ & $22.04$ & $0.144$ & $20.67$ & $0.162$ \\
C2F-LRTC-TV-II & $\textcolor{blue}{\mathbf{25.37}}$ & $\textcolor{blue}{\mathbf{0.088}}$ & $\textcolor{blue}{\mathbf{25.51}}$ & $\textcolor{blue}{\mathbf{0.095}}$ & $\textcolor{blue}{\mathbf{23.46}}$ & $\textcolor{blue}{\mathbf{0.122}}$ & $\textcolor{blue}{\mathbf{21.61}}$ & $\textcolor{blue}{\mathbf{0.146}}$ \\
Short-Cut C2F & $22.34$ & $0.125$ & $23.69$ & $0.118$ & $20.57$ & $0.171$ & $19.72$ & $0.181$ \\
\bottomrule
\end{tabular}
}
\end{table}

\begin{table}[t]
\scriptsize
\centering
\caption{The average relative patch rank (RPR) for patches of different sizes in the ground truth. In each column, the average RPR decreases as the patch size becomes smaller, which is consistent with our experimental settings.}
\label{tab:rank_analysis}
\setlength{\tabcolsep}{0.8mm}{
\begin{tabular}{lcccccccc}
\toprule
~ &  \textbf{Airplane} & \textbf{Baboon} & \textbf{Barbara} & \textbf{Facade} & \textbf{House} & \textbf{Lena} & \textbf{Peppers} & \textbf{Sailboat}\\
\midrule
\textbf{Coarse} & $0.2654$ & $0.4889$ & $0.3359$ & $0.3190$ & $0.2405$ & $0.3072$ & $0.2954$ & $0.3647$\\
\textbf{Fine-1} & $0.2053$ & $0.4366$ & $0.2953$ & $0.2651$ & $0.2071$ & $0.2572$ & $0.2621$ & $0.3068$ \\
\textbf{Fine-2} & $0.1621$ & $0.4141$ & $0.2615$ & $0.2266$ & $0.1723$ & $0.2249$ & $0.2286$ & $0.2801$ \\
\textbf{Fine-3} & $0.1415$ & $0.3867$ & $0.2289$ & $0.2004$ & $0.1561$ & $0.1978$ & $0.2078$ & $0.2529$\\
\bottomrule
\end{tabular}
}
\vspace{-3mm}
\end{table}

\begin{table*}[t]
\scriptsize
\centering
\caption{The restoration performance of pure LRTC and C2F-LRTC methods under different missing ratios.}
\label{tab:results_sum}
\setlength{\tabcolsep}{2.0mm}{
\begin{tabular}{l|cc|cc|cc||l|cc|cc|cc}
\toprule
\multirow{2}{*}{\footnotesize{\textbf{Airplane}}} & \multicolumn{2}{c|}{\textbf{0.7}} & \multicolumn{2}{c|}{\textbf{0.8}} & \multicolumn{2}{c||}{\textbf{0.9}} & \multirow{2}{*}{\footnotesize{\textbf{Baboon}}} & \multicolumn{2}{c|}{\textbf{0.7}} & \multicolumn{2}{c|}{\textbf{0.8}} & \multicolumn{2}{c}{\textbf{0.9}} \\
~ & \textbf{PSNR} & \textbf{RSE} & \textbf{PSNR} & \textbf{RSE} & \textbf{PSNR} & \textbf{RSE} & ~ & \textbf{PSNR} & \textbf{RSE} & \textbf{PSNR} & \textbf{RSE} & \textbf{PSNR} & \textbf{RSE} \\
\midrule
HaLRTC & $24.50$ & $0.074$ & $21.97$ & $0.099$ & $18.97$ & $0.140$ & HaLRTC & $21.94$ & $0.148$ & $20.48$ & $0.175$ & $18.60$ & $0.220$ \\  
C2F-HaLRTC & $\textcolor{blue}{\mathbf{24.76}}$ & $\textcolor{blue}{\mathbf{0.072}}$ & $\textcolor{blue}{\mathbf{22.26}}$ & $\textcolor{blue}{\mathbf{0.096}}$ & $\textcolor{blue}{\mathbf{19.45}}$ & $\textcolor{blue}{\mathbf{0.133}}$ & C2F-HaLRTC & $\textcolor{blue}{\mathbf{22.16}}$ & $\textcolor{blue}{\mathbf{0.144}}$ & $\textcolor{blue}{\mathbf{20.69}}$ & $\textcolor{blue}{\mathbf{0.171}}$ & $\textcolor{blue}{\mathbf{18.62}}$ & $\textcolor{blue}{\mathbf{0.217}}$ \\ 
STDC & $22.77$ & $0.090$ & $18.67$ & $0.145$ & $15.25$ & $0.236$ & STDC & $17.19$ & $0.256$ & $16.50$ & $0.277$ & $14.48$ & $0.350$ \\
C2F-STDC & $\textcolor{blue}{\mathbf{23.19}}$ & $\textcolor{blue}{\mathbf{0.086}}$ & $\textcolor{blue}{\mathbf{21.16}}$ & $\textcolor{blue}{\mathbf{0.137}}$ & $\textcolor{blue}{\mathbf{20.13}}$ & $\textcolor{blue}{\mathbf{0.123}}$ & C2F-STDC & $\textcolor{blue}{\mathbf{20.19}}$ & $\textcolor{blue}{\mathbf{0.176}}$ & $\textcolor{blue}{\mathbf{19.50}}$ & $\textcolor{blue}{\mathbf{0.197}}$ & $\textcolor{blue}{\mathbf{17.19}}$ & $\textcolor{blue}{\mathbf{0.256}}$ \\ 
LRTC-TV-II & $26.91$ & $0.056$ & $27.33$ & $0.068$ & $22.80$ & $0.098$ & LRTC-TV-II & $23.30$ & $0.129$ & $22.32$ & $0.143$ & $21.22$ & $0.165$ \\
C2F-LRTC-TV-II & $\textcolor{blue}{\mathbf{27.84}}$ & $\textcolor{blue}{\mathbf{0.051}}$ & $\textcolor{blue}{\mathbf{27.35}}$ & $\textcolor{blue}{\mathbf{0.067}}$ & $\textcolor{blue}{\mathbf{23.28}}$ & $\textcolor{blue}{\mathbf{0.085}}$ & C2F-LRTC-TV-II & $\textcolor{blue}{\mathbf{23.47}}$ & $\textcolor{blue}{\mathbf{0.126}}$ & $\textcolor{blue}{\mathbf{22.52}}$ & $\textcolor{blue}{\mathbf{0.141}}$ & $\textcolor{blue}{\mathbf{23.05}}$ & $\textcolor{blue}{\mathbf{0.161}}$ \\ 
LRTC-PDS & $25.42$ & $0.072$ & $23.55$ & $0.089$ & $20.72$ & $0.144$ & LRTC-PDS & $23.17$ & $0.136$ & $22.07$ & $0.154$ & $20.72$ & $0.245$ \\
C2F-LRTC-PDS & $\textcolor{blue}{\mathbf{25.79}}$ & $\textcolor{blue}{\mathbf{0.066}}$ & $\textcolor{blue}{\mathbf{23.83}}$ & $\textcolor{blue}{\mathbf{0.084}}$ & $\textcolor{blue}{\mathbf{21.15}}$ & $\textcolor{blue}{\mathbf{0.140}}$ & C2F-LRTC-PDS & $\textcolor{blue}{\mathbf{23.25}}$ & $\textcolor{blue}{\mathbf{0.131}}$ & $\textcolor{blue}{\mathbf{22.15}}$ & $\textcolor{blue}{\mathbf{0.150}}$ & $\textcolor{blue}{\mathbf{20.75}}$ & $\textcolor{blue}{\mathbf{0.239}}$ \\
\midrule
\multirow{2}{*}{\footnotesize{\textbf{Barbara}}} & \multicolumn{2}{c|}{\textbf{0.7}} & \multicolumn{2}{c|}{\textbf{0.8}} & \multicolumn{2}{c||}{\textbf{0.9}} & \multirow{2}{*}{\footnotesize{\textbf{Facade}}} & \multicolumn{2}{c|}{\textbf{0.7}} & \multicolumn{2}{c|}{\textbf{0.8}} & \multicolumn{2}{c}{\textbf{0.9}} \\
~ & \textbf{PSNR} & \textbf{RSE} & \textbf{PSNR} & \textbf{RSE} & \textbf{PSNR} & \textbf{RSE} & ~ & \textbf{PSNR} & \textbf{RSE} & \textbf{PSNR} & \textbf{RSE} & \textbf{PSNR} & \textbf{RSE} \\
\midrule
HaLRTC & $25.27$ & $0.112$ & $22.65$ & $0.151$ & $19.06$ & $0.229$ & HaLRTC & $28.58$ & $0.083$ & $26.09$ & $0.110$ & $22.57$ & $0.144$ \\  
C2F-HaLRTC & $\textcolor{blue}{\mathbf{25.72}}$ & $\textcolor{blue}{\mathbf{0.106}}$ & $\textcolor{blue}{\mathbf{22.87}}$ & $\textcolor{blue}{\mathbf{0.147}}$ & $\textcolor{blue}{\mathbf{19.26}}$ & $\textcolor{blue}{\mathbf{0.223}}$ & C2F-HaLRTC & $\textcolor{blue}{\mathbf{29.74}}$ & $\textcolor{blue}{\mathbf{0.063}}$ & $\textcolor{blue}{\mathbf{27.54}}$ & $\textcolor{blue}{\mathbf{0.081}}$ & $\textcolor{blue}{\mathbf{24.87}}$ & $\textcolor{blue}{\mathbf{0.111}}$ \\ 
STDC & $22.04$ & $0.162$ & $19.84$ & $0.209$ & $16.28$ & $0.315$ & STDC & $26.54$ & $0.091$ & $24.54$ & $0.115$ & $20.95$ & $0.153$ \\
C2F-STDC & $\textcolor{blue}{\mathbf{22.17}}$ & $\textcolor{blue}{\mathbf{0.160}}$ & $\textcolor{blue}{\mathbf{20.90}}$ & $\textcolor{blue}{\mathbf{0.207}}$ & $\textcolor{blue}{\mathbf{20.41}}$ & $\textcolor{blue}{\mathbf{0.196}}$ & C2F-STDC & $\textcolor{blue}{\mathbf{26.64}}$ & $\textcolor{blue}{\mathbf{0.090}}$ & $\textcolor{blue}{\mathbf{24.56}}$ & $\textcolor{blue}{\mathbf{0.114}}$ & $\textcolor{blue}{\mathbf{22.62}}$ & $\textcolor{blue}{\mathbf{0.143}}$ \\ 
LRTC-TV-II & $27.88$ & $0.085$ & $26.06$ & $0.100$ & $23.87$ & $0.131$ & LRTC-TV-II & $27.19$ & $0.092$ & $25.76$ & $0.099$ & $21.79$ & $0.158$ \\
C2F-LRTC-TV-II & $\textcolor{blue}{\mathbf{28.46}}$ & $\textcolor{blue}{\mathbf{0.078}}$ & $\textcolor{blue}{\mathbf{26.69}}$ & $\textcolor{blue}{\mathbf{0.097}}$ & $\textcolor{blue}{\mathbf{24.75}}$ & $\textcolor{blue}{\mathbf{0.119}}$ & C2F-LRTC-TV-II & $\textcolor{blue}{\mathbf{27.89}}$ & $\textcolor{blue}{\mathbf{0.068}}$ & $\textcolor{blue}{\mathbf{25.80}}$ & $\textcolor{blue}{\mathbf{0.091}}$ & $\textcolor{blue}{\mathbf{24.40}}$ & $\textcolor{blue}{\mathbf{0.117}}$ \\ 
LRTC-PDS & $26.86$ & $0.107$ & $25.13$ & $0.132$ & $22.13$ & $0.185$ & LRTC-PDS & $24.01$ & $0.130$ & $22.21$ & $0.159$ & $19.93$ & $0.246$ \\
C2F-LRTC-PDS & $\textcolor{blue}{\mathbf{27.00}}$ & $\textcolor{blue}{\mathbf{0.098}}$ & $\textcolor{blue}{\mathbf{25.38}}$ & $\textcolor{blue}{\mathbf{0.124}}$ & $\textcolor{blue}{\mathbf{22.34}}$ & $\textcolor{blue}{\mathbf{0.178}}$ & C2F-LRTC-PDS & $\textcolor{blue}{\mathbf{24.41}}$ & $\textcolor{blue}{\mathbf{0.123}}$ & $\textcolor{blue}{\mathbf{22.58}}$ & $\textcolor{blue}{\mathbf{0.153}}$ & $\textcolor{blue}{\mathbf{20.09}}$ & $\textcolor{blue}{\mathbf{0.239}}$ \\
\midrule
\multirow{2}{*}{\footnotesize{\textbf{House}}} & \multicolumn{2}{c|}{\textbf{0.7}} & \multicolumn{2}{c|}{\textbf{0.8}} & \multicolumn{2}{c||}{\textbf{0.9}} & \multirow{2}{*}{\footnotesize{\textbf{Lena}}} & \multicolumn{2}{c|}{\textbf{0.7}} & \multicolumn{2}{c|}{\textbf{0.8}} & \multicolumn{2}{c}{\textbf{0.9}} \\
~ & \textbf{PSNR} & \textbf{RSE} & \textbf{PSNR} & \textbf{RSE} & \textbf{PSNR} & \textbf{RSE} & ~ & \textbf{PSNR} & \textbf{RSE} & \textbf{PSNR} & \textbf{RSE} & \textbf{PSNR} & \textbf{RSE} \\
\midrule
HaLRTC & $25.79$ & $0.084$ & $23.72$ & $0.107$ & $20.54$ & $0.154$ & HaLRTC & $25.86$ & $0.093$ & $23.19$ & $0.125$ & $19.77$ & $0.185$ \\  
C2F-HaLRTC & $\textcolor{blue}{\mathbf{25.92}}$ & $\textcolor{blue}{\mathbf{0.830}}$ & $\textcolor{blue}{\mathbf{24.07}}$ & $\textcolor{blue}{\mathbf{0.103}}$ & $\textcolor{blue}{\mathbf{21.05}}$ & $\textcolor{blue}{\mathbf{0.145}}$ & C2F-HaLRTC & $\textcolor{blue}{\mathbf{25.99}}$ & $\textcolor{blue}{\mathbf{0.091}}$ & $\textcolor{blue}{\mathbf{23.28}}$ & $\textcolor{blue}{\mathbf{0.124}}$ & $\textcolor{blue}{\mathbf{20.15}}$ & $\textcolor{blue}{\mathbf{0.177}}$ \\ 
STDC & $24.28$ & $0.100$ & $22.50$ & $0.123$ & $17.54$ & $0.218$ & STDC & $23.22$ & $0.125$ & $20.45$ & $0.171$ & $16.64$ & $0.266$ \\
C2F-STDC & $\textcolor{blue}{\mathbf{24.73}}$ & $\textcolor{blue}{\mathbf{0.095}}$ & $\textcolor{blue}{\mathbf{22.88}}$ & $\textcolor{blue}{\mathbf{0.118}}$ & $\textcolor{blue}{\mathbf{21.91}}$ & $\textcolor{blue}{\mathbf{0.132}}$ & C2F-STDC & $\textcolor{blue}{\mathbf{22.64}}$ & $\textcolor{blue}{\mathbf{0.118}}$ & $\textcolor{blue}{\mathbf{21.56}}$ & $\textcolor{blue}{\mathbf{0.169}}$ & $\textcolor{blue}{\mathbf{20.84}}$ & $\textcolor{blue}{\mathbf{0.164}}$ \\
LRTC-TV-II & $28.27$ & $0.065$ & $27.37$ & $0.074$& $24.99$ & $0.094$ & LRTC-TV-II & $29.62$ & $0.059$ & $27.52$ & $0.072$ & $25.36$ & $0.097$ \\
C2F-LRTC-TV-II & $\textcolor{blue}{\mathbf{28.34}}$ & $\textcolor{blue}{\mathbf{0.064}}$ & $\textcolor{blue}{\mathbf{27.40}}$ & $\textcolor{blue}{\mathbf{0.073}}$ & $\textcolor{blue}{\mathbf{25.37}}$ & $\textcolor{blue}{\mathbf{0.088}}$ & C2F-LRTC-TV-II & $\textcolor{blue}{\mathbf{30.04}}$ & $\textcolor{blue}{\mathbf{0.058}}$ & $\textcolor{blue}{\mathbf{27.92}}$ & $\textcolor{blue}{\mathbf{0.070}}$ & $\textcolor{blue}{\mathbf{25.51}}$ & $\textcolor{blue}{\mathbf{0.095}}$ \\ 
LRTC-PDS & $27.19$ & $0.075$ & $25.63$ & $0.096$ & $22.77$ & $0.146$ & LRTC-PDS & $28.48$ & $0.086$ & $26.39$ & $0.106$ & $22.40$ & $0.141$ \\
C2F-LRTC-PDS & $\textcolor{blue}{\mathbf{27.30}}$ & $\textcolor{blue}{\mathbf{0.066}}$ & $\textcolor{blue}{\mathbf{25.74}}$ & $\textcolor{blue}{\mathbf{0.089}}$ & $\textcolor{blue}{\mathbf{22.85}}$ & $\textcolor{blue}{\mathbf{0.142}}$ & C2F-LRTC-PDS & $\textcolor{blue}{\mathbf{28.55}}$ & $\textcolor{blue}{\mathbf{0.077}}$ & $\textcolor{blue}{\mathbf{26.49}}$ & $\textcolor{blue}{\mathbf{0.099}}$ & $\textcolor{blue}{\mathbf{22.46}}$ & $\textcolor{blue}{\mathbf{0.135}}$ \\
\midrule
\multirow{2}{*}{\footnotesize{\textbf{Peppers}}} &  \multicolumn{2}{c|}{\textbf{0.7}} & \multicolumn{2}{c|}{\textbf{0.8}} & \multicolumn{2}{c||}{\textbf{0.9}} & \multirow{2}{*}{\footnotesize{\textbf{Sailboat}}} &  \multicolumn{2}{c|}{\textbf{0.7}} & \multicolumn{2}{c|}{\textbf{0.8}} & \multicolumn{2}{c}{\textbf{0.9}} \\
~ & \textbf{PSNR} & \textbf{RSE} & \textbf{PSNR} & \textbf{RSE} & \textbf{PSNR} & \textbf{RSE} & ~ & \textbf{PSNR} & \textbf{RSE} & \textbf{PSNR} & \textbf{RSE} & \textbf{PSNR} & \textbf{RSE} \\
\midrule
HaLRTC & $22.82$ & $0.132$ & $20.31$ & $0.176$ & $16.86$ & $0.262$ & HaLRTC & $22.78$ & $0.127$ & $20.60$ & $0.164$ & $17.81$ & $0.226$ \\  
C2F-HaLRTC & $\textcolor{blue}{\mathbf{23.59}}$ & $\textcolor{blue}{\mathbf{0.121}}$ & $\textcolor{blue}{\mathbf{20.72}}$ & $\textcolor{blue}{\mathbf{0.168}}$ & $\textcolor{blue}{\mathbf{16.86}}$ & $\textcolor{blue}{\mathbf{0.261}}$ & C2F-HaLRTC & $\textcolor{blue}{\mathbf{23.09}}$ & $\textcolor{blue}{\mathbf{0.123}}$ & $\textcolor{blue}{\mathbf{20.87}}$ & $\textcolor{blue}{\mathbf{0.159}}$ & $\textcolor{blue}{\mathbf{17.93}}$ & $\textcolor{blue}{\mathbf{0.223}}$ \\ 
STDC & $21.37$ & $0.156$ & $18.61$ & $0.214$ & $16.31$ & $0.347$ & STDC & $20.58$ & $0.184$ & $18.10$ & $0.245$ & $14.87$ & $0.317$ \\
C2F-STDC & $\textcolor{blue}{\mathbf{21.48}}$ & $\textcolor{blue}{\mathbf{0.154}}$ & $\textcolor{blue}{\mathbf{20.64}}$ & $\textcolor{blue}{\mathbf{0.210}}$ & $\textcolor{blue}{\mathbf{18.42}}$ & $\textcolor{blue}{\mathbf{0.219}}$ & C2F-STDC & $\textcolor{blue}{\mathbf{20.85}}$ & $\textcolor{blue}{\mathbf{0.178}}$ & $\textcolor{blue}{\mathbf{18.23}}$ & $\textcolor{blue}{\mathbf{0.185}}$ & $\textcolor{blue}{\mathbf{19.12}}$ & $\textcolor{blue}{\mathbf{0.194}}$ \\
LRTC-TV-II & $27.75$ & $0.076$ & $25.25$ & $0.103$ & $22.04$ & $0.144$ & LRTC-TV-II & $25.21$ & $0.103$ & $23.42$ & $0.118$ & $20.67$ & $0.162$ \\
C2F-LRTC-TV-II & $\textcolor{blue}{\mathbf{28.37}}$ & $\textcolor{blue}{\mathbf{0.072}}$ & $\textcolor{blue}{\mathbf{26.17}}$ & $\textcolor{blue}{\mathbf{0.096}}$ & $\textcolor{blue}{\mathbf{23.46}}$ & $\textcolor{blue}{\mathbf{0.122}}$ & C2F-LRTC-TV-II & $\textcolor{blue}{\mathbf{26.61}}$ & $\textcolor{blue}{\mathbf{0.092}}$ & $\textcolor{blue}{\mathbf{23.76}}$ & $\textcolor{blue}{\mathbf{0.112}}$ & $\textcolor{blue}{\mathbf{21.61}}$ & $\textcolor{blue}{\mathbf{0.146}}$ \\
LRTC-PDS & $27.05$ & $0.102$ & $24.06$ & $0.131$ & $20.09$ & $0.194$ & LRTC-PDS & $23.95$ & $0.123$ & $22.13$ & $0.153$ & $19.37$ & $0.209$ \\
C2F-LRTC-PDS & $\textcolor{blue}{\mathbf{27.33}}$ & $\textcolor{blue}{\mathbf{0.092}}$ & $\textcolor{blue}{\mathbf{24.31}}$ & $\textcolor{blue}{\mathbf{0.122}}$ & $\textcolor{blue}{\mathbf{20.35}}$ & $\textcolor{blue}{\mathbf{0.188}}$ & C2F-LRTC-PDS & $\textcolor{blue}{\mathbf{24.33}}$ & $\textcolor{blue}{\mathbf{0.117}}$ & $\textcolor{blue}{\mathbf{22.46}}$ & $\textcolor{blue}{\mathbf{0.147}}$ & $\textcolor{blue}{\mathbf{19.58}}$ & $\textcolor{blue}{\mathbf{0.205}}$ \\
\bottomrule
\end{tabular}}
\end{table*}

\subsection{Pure LRTC Compared with C2F-LRTC}
\label{exp:pure_vs_C2F}

To evaluate the performance of image completion, we employ the RSE and peak signal-to-noise ratio (PSNR) as the evaluation metrics, which are widely used in image restoration tasks. The definition of those two metrics are as follows:
\begin{equation}
\small{
\begin{aligned}
\rm{RSE} &= \frac{|| \ten{Z} - \ten{Z}_{true} ||_F}{|| \ten{Z}_{true} ||_F}\\ \rm{PSNR} &= 10\log_{10} \frac{\prod_{k=1}^{d} \hat{\ten{Z}}^2_{true}}{|| \ten{Z} - \ten{Z}_{true} ||_F^2}    
\end{aligned}  
}
\end{equation}
where $\ten{Z}$, $\ten{Z}_{true}$ and $\hat{\ten{Z}}_{true}$ represent the completed image, ground truth image, and the maximum value of the pixels in the ground truth image, respectively. The performance of the restoration method is inversely proportional to the value of RSE and proportional to the value of PSNR.

For the C2F implementation, we set the number of fine stage repetitions $F=3$, and the initial patch replace threshold $\epsilon=0.15$. For the second and the third fine stages, the replacement threshold will be updated as below: 
\begin{equation}
\label{eq:epsilon}
\small{
\epsilon=\left\{
\begin{array}{l}
\frac{3}{2}\max\left(\rm{gap}(\hat{\ten{Z}}^{(k)}_{temp}, \ten{Z}^{(k)}_{temp})\right)\text{, if $F=2$,} \\
\frac{3}{2}\cdot \max\left(\rm{gap}(\hat{\ten{Z}}^{(k)}_{temp}, \ten{Z}^{(k)}_{temp})\mathbf{r}_k\right)\text{, if $F=3$,} \\
\end{array}
\right.
}
\end{equation}
where the vector $\mathbf{r}$ is binary with elements $0$ or $1$, recording whether a completed patch in the second fine stage replaces its counterpart in the coarse stage or not. We remark that the threshold becomes larger as the patch size decreases.

Table~\ref{tab:results_sum} summarizes the comparison results of the pure LRTC and their C2F version under three different missing ratios (viz. $0.7$, $0.8$, and $0.9$). For easy reading, the C2F-LRTC results are marked blue, and are put below the corresponding pure LRTC method. It can be observed that under all missing ratios, the performance of \textbf{\textit{every}} pure LRTC method is enhanced after employing the C2F strategy. In other words, C2F-LRTC can restore images with larger PSNR and smaller RSE values. Furthermore, it can be seen that as the missing ratio increases, C2F has a trend to boost the performance of pure LRTC methods more significantly. For example, when completing the image Barbara, the gaps between the PSNR of STDC and C2F-STDC are $0.13$, $1.06$, and $4.13$ as the missing ratio increases. To sense the improvement brought by C2F more intuitively, Fig.~\ref{fig:facade_example} shows the restoration results of facade and sailboat under $90\%$ missing ratio. These two images are representative, since the image facade has \textbf{\textit{regular patterns}} while the image sailboat contains \textbf{\textit{both}} the \textbf{\textit{details-lacking parts}} (e.g., the sky and the lake) and the \textbf{\textit{complex objects}} (e.g., the trees and the boat). It is noticeable that C2F-LRTC restores both two kinds of images with richer details. In summary, C2F can steadily improve the performance of existing LRTC methods and retain more details.

\subsection{The Effectiveness of Successive Fine-grained Completion}
\label{exp:gradual_refine}

In this section, we validate the effectiveness of the successive fine-grained completion by comparing the restoration results obtained by pure LRTC, C2F-LRTC, and the Short-Cut C2F-LRTC, which only contains the coarse stage at the beginning and the last fine stage with the smallest patches.

The results are displayed in Table~\ref{tab:short_cut}, where we employ the LRTC-TV-II algorithm and set the missing ratio as $90\%$. For the Short-Cut C2F, we set a larger patch replace threshold $\epsilon = 0.3$ to ensure that the newly completed patches in the last fine stage can replace their counterparts in the coarse stage. For the remaining hyper-parameters, we keep them the same as in the original C2F. It is seen that instead of improving the performance of pure LRTC, the Short-Cut C2F \textit{\textbf{degrades}} the restoration ability of the pure LRTC, which demonstrates the importance of the successive refinement in our C2F procedure.

\subsection{Local Rank Analysis}
\label{exp:local_rank}
We then validate the correctness of our assumption that smaller local ranks go along with decreasing patch sizes. We compute the average relative patch ranks (RPR) for patches of different sizes in the ground truth. The RPR is defined as the ratio between the number of singular values (SVs), which accounts for $90\%$ of the total singular values summation, and the patch size. Along this side, the patch size can be ignored when comparing the patch ranks. The average RPR in the coarse and three fine stages are listed in Table~\ref{tab:rank_analysis}. For all eight images, it is seen that the smaller the patch size, the smaller RPR, which we adopt as a general rule applicable to most real-life images. We emphasize that the decreasing local ranks in the fine stages will not degrade the global rank of the restored image. For instance, the global ranks (number of SVs that takes up $90\%$ of the summation of the total SVs) of the image facade restored by the pure LRTC-TV-II and C2F-LRTC-TV-II in Fig.~\ref{fig:facade_example} are $49$ and $50$, respectively. Therefore, we conclude that it is insufficient to set a global rank only to complete partially observed images, and appropriate local ranks are important to restore local structures. 

\section{Conclusion}
\label{sec:conclusion}
This work presents a general and easily implementable coarse-to-fine framework called C2F, which can effectively boost the performance of any existing LRTC approach with the mindfully designed progressive refinement process and appropriate local ranks. To the best of our knowledge, it is the first time that global and local ranks are both utilized for completing images. By employing the proposed C2F strategy, both the global and local tensor structures are utilized, leading to a higher completion performance. Extensive experiments are done to demonstrate the practicality of C2F, which sheds light on boosting images restoration results and LRTC techniques performance in a divide-and-conquer manner.

\bibliographystyle{IEEEtran}
\bibliography{egbib}

\end{document}